\documentclass[preprint,12pt,authoryear]{elsarticle}

\usepackage{amssymb}
\usepackage{amsmath}
\usepackage{algorithm}
\usepackage{algpseudocode}
\usepackage{hyperref}
\usepackage{algorithm}
\usepackage{tikz}
\usetikzlibrary{shapes, arrows.meta, positioning}
\usepackage{graphicx}
\usepackage{tabularx}
\usepackage{subfigure}
\usepackage{booktabs}

\journal{Computers \& Chemical Engineering}

\begin{document}

\begin{frontmatter}



\title{\emph{FaultExplainer}: Leveraging Large Language Models for Interpretable Fault Detection and Diagnosis}

\author[purduecheme]{Abdullah Khan}
\author[purduecheme]{Rahul Nahar}
\author[purduecheme]{Hao Chen}

\author[purduecheme]{Gonzalo E. Constante Flores}

\author[purduecheme]{Can Li\corref{cor}}
\ead{canli@purdue.edu}
\cortext[cor]{Corresponding author at: Davidson School of Chemical Engineering, Purdue University, USA.}

\affiliation[purduecheme]{organization={Davidson School of Chemical Engineering, Purdue University},
            addressline={480 W. Stadium Ave}, 
            city={West Lafayette},
            postcode={47907}, 
            state={IN},
            country={USA}}

\begin{abstract}
Machine learning algorithms are increasingly being applied to fault detection and diagnosis (FDD) in chemical processes. However, existing data-driven FDD platforms often lack interpretability for process operators and struggle to identify root causes of previously unseen faults. This paper presents \emph{FaultExplainer}, an interactive tool designed to improve fault detection, diagnosis, and explanation in the Tennessee Eastman Process (TEP). \emph{FaultExplainer} integrates real-time sensor data visualization, Principal Component Analysis (PCA)-based fault detection, and identification of top contributing variables within an interactive user interface powered by large language models (LLMs). We evaluate the LLMs' reasoning capabilities in two scenarios: one where historical root causes are provided, and one where they are not to mimic the challenge of previously unseen faults.  Experimental results using GPT-4o and o1-preview models demonstrate the system's strengths in generating plausible and actionable explanations, while also highlighting its limitations, including reliance on PCA-selected features and occasional hallucinations.
\end{abstract}


\begin{highlights}
    \item \emph{FaultExplainer} is an LLM-based natural language system for fault detection, diagnosis, and explanation.
    \item \emph{FaultExplainer} integrates PCA and $T^2$ statistics with process decriptions for grounded fault explanations.
    \item \emph{FaultExplainer} provides a web interface for real-time monitoring and user-friendly interaction.
    \item Results show \emph{FaultExplainer} has reasonable performance in diagnosing unseen faults.
\end{highlights}

\begin{keyword}
Fault detection and diagnosis \sep Large language models\sep Explainable AI
\end{keyword}

\end{frontmatter}


\section{Introduction}
Fault Detection and Diagnosis (FDD) are critical components in ensuring the reliability and safety of chemical processes. Fault detection involves recognizing the occurrence of a fault in the system, while fault diagnosis ascertains the root cause of the fault. In the chemical industry, where processes are complex and highly integrated, early and accurate FDD can prevent significant economic losses, environmental harm, and safety hazards. Effective FDD systems enhance operational efficiency by minimizing downtime and maintenance costs, thereby ensuring the continuous and safe operation of chemical plants.

Various approaches have been developed for FDD, broadly categorized into model-based and data-driven methods \citep{Chiang2001}. Model-based methods rely on the expertise and experience of operators and engineers and physics-based equations, often implemented through expert systems \citep{Venkatasubramanian2003I,Venkatasubramanian2003II}. These methods can be highly effective in systems where physical models are well understood and can provide detailed explanations of fault mechanisms. However, they require extensive domain knowledge and may not adapt well to changes in the process.

Data-driven methods, on the other hand, utilize historical and real-time process data to detect and diagnose faults \citep{Venkatasubramanian2003III}. Traditional statistical methods such as Principal Component Analysis (PCA) have been widely used for FDD due to their ability to handle multivariate data. However, these methods assume linearity and Gaussian distribution of data, which limits their applicability to more complex, non-linear processes commonly found in chemical industries. Recent advances in machine learning, particularly deep learning, have introduced more robust data-driven FDD techniques. Deep learning methods, such as  Recurrent Neural Networks (RNNs), autoencoders, deep belief networks, and variants \citep{Sun2020,zhang2022dynamic,zhang2017deep}, can model complex non-linear relationships and temporal dependencies in process data. 

While deep learning methods offer superior modeling capabilities, they also come with challenges. One of the primary disadvantages is their lack of interpretability; understanding the reasoning behind a model's predictions can be difficult, which poses a challenge to gaining trust in these systems. To make the deep learning models interpretable for FDD, recent works by \cite{bhakte2022explainable,bhakte2023alarm} have leveraged techniques in explainable artificial intelligence (XAI) such as Shapley value \citep{hart1989shapley} to attribute the predictions of deep learning model to the contributions of each process variable. These XAI approaches are more interpretable than the deep learning models alone but they can only provide quantitative metrics and statistics of the models.

To make FDD more interpretable, ideally, one would prefer to develop a natural-language-based explanation and query systems such that the process operators can have more intuitive interpretations of monitoring data using their domain knowledge. Such a natural language-based system for FDD was not possible until the recent advances in large language models (LLMs) such as GPT-4 \citep{OpenAI2023}  and Llama-2 \citep{Touvron2023LlamaModels}. However, the limitation of the LLMs is that they tend to ``hallucinate'', meaning that they could generate text that is incorrect, nonsensical, or not real. This is problematic for a safety-critical application like FDD. 

Another limitation of data-driven fault diagnosis methods is their inability to diagnose the root causes of faults that have never been encountered. In practice, data-driven models often classify unseen faults as ``unknown'' without providing meaningful explanations\citep{Chiang2001}. Addressing these novel scenarios requires domain knowledge and potentially model-based approaches. Whether LLMs can leverage observed variations in measured and manipulated variables to infer causes of unseen faults remains an open question.

To address these challenges, we develop \emph{FaultExplainer}, a first-of-its-kind natural language-based system for intuitive fault detection and diagnosis (FDD) in chemical plants. In this work, we demonstrate a novel integration of LLMs with traditional statistical learning methods to generate explanations and root cause analyses for faults detected in the Tennessee Eastman Process (TEP) \citep{downs1993plant}. To mitigate the hallucinations of LLMs, we propose a method that leverages their capabilities while grounding their explanations in a detailed TEP process description, PCA, and feature importance analysis, ensuring that the resulting explanations are both accurate and interpretable for chemical process engineers. We evaluate the LLMs' reasoning capabilities in two scenarios: one where historical root causes are provided, and one where they are not to mimic the challenge of previously unseen faults. \emph{FaultExplainer} is released as an open-source package with video demonstrations, available at \url{https://github.com/li-group/FaultExplainer}.

The rest of the paper is organized as follows: Section \ref{sec:background} presents the necessary background and related literature. Section \ref{sec:overview} provides an overview of \emph{FaultExplainer}'s capabilities. The methodologies are detailed in Section \ref{sec:methods}. Section \ref{sec:results} evaluates \emph{FaultExplainer} using the 15 known faults from the TEP process. Finally, conclusions are drawn in Section \ref{sec:conclusions}.

\section{Background and Literature Review}\label{sec:background}

Fault Detection and Diagnosis (FDD) methods can be broadly classified into two main categories: model-based and data-driven approaches. Model-based methods rely on physical and chemical models of the process and can be further divided into quantitative and qualitative methods \citep{Venkatasubramanian2003I,Venkatasubramanian2003II}. Quantitative methods use mathematical models to represent the process. Diagnostic observers \citep{Watanabe1983I,watanabe1983fault} are algorithms that compare the actual process outputs with the outputs of a mathematical model to detect discrepancies that indicate faults. Parity relations \citep{gertler1995generating,Yin2017} generate residuals by comparing measurements to their estimated values based on mathematical models, and faults are detected when these residuals exceed a certain threshold. Kalman filters \citep{welch1995introduction} estimate the state of a process in real-time and use the discrepancies between the predicted and actual states to identify faults. Parameter estimation \citep{jiang2008parameter,Alanqar2017} involves estimating the parameters of a process model, where significant deviations from normal parameter values indicate potential faults. 

Qualitative methods use non-numeric models to describe the process. Causal models including Digraphs \citep{Vedam1997}, fault trees \citep{geymayr1995fault}, and qualitative physics models \citep{sacks1988qualitative}, have been used to represent and analyze the cause-effect relationships within a process. Digraphs, or directed graphs, visually depict the cause-effect relationships between variables in a system, allowing for the identification of fault propagation paths. Fault trees utilize logic gates to model the sequences of events and conditions that can lead to system failures. Qualitative physics models use qualitative reasoning to represent the physical and functional relationships within a system to facilitate the understanding and diagnosis of faults based on causal interactions. 

Data-driven methods utilize statistical techniques and machine learning algorithms to identify patterns and anomalies in process data \citep{Venkatasubramanian2003III}. Principal Component Analysis (PCA) \citep{choi2005fault} reduces the dimensionality of process data by identifying the principal components that capture the most variance, helping in detecting anomalies by highlighting deviations from the normal data patterns. Partial Least Squares (PLS) \citep{jia2016quality} is used for modeling relationships between input and output variables in a process and can detect faults by identifying deviations in these relationships. Deep learning models including RNNs \citep{Sun2020}, autoencoders \citep{zhang2022dynamic}, deep belief networks \citep{zhang2017deep}, and Bayesian networks \citep{Kumari2022I,Kumari2022II}, are recently used for fault detection. Despite their effectiveness, data-driven methods often suffer from a lack of interpretability, making it difficult for domain experts to understand and trust the results.

Explainable artificial intelligence (XAI) is an area that has attracted significant attention in the machine learning community, aiming to improve the interpretability of deep learning models. Broadly speaking, the current XAI methods can be classified into two main categories \citep{Murdoch2019DefinitionsLearning, Mittelstadt2019ExplainingAI}. Model-based approaches construct machine learning models with the goal of explainability in mind, resulting in simple, easier-to-interpret models. For example, one can use straightforward models such as linear/logistic regression \citep{Bursac2008PurposefulRegression}, decision trees \citep{Quinlan1987SimplifyingTrees}, and K-nearest neighbor \citep{Peterson2009K-nearestNeighbor}. On the other hand, post hoc analysis refers to interpreting the model after it has been trained. This is usually applied to black-box machine learning models, such as neural networks, where the practitioner analyzes a trained model to provide insights into the learned relationships. An example of post hoc analysis is the linear approximation of the black-box model around a reference point, allowing the weights of the linear model to be directly interpreted as sensitivity measures. This indicates how much the classifier will respond as a particular feature changes. Examples of linear approximation include LIME \citep{Ribeiro2016WhyClassifier}. Another popular local approach is contrastive/counterfactual explanation (CFE) \citep{Wachter2017CounterfactualGDPR}, where the idea is to explain to the practitioner how the features would have to change for the model to predict an alternative outcome. Recently, XAI methods have been applied to fault detection and diagnosis (FDD) applications. For instance, \cite{Hale2022} use symbolic regression to generate explainable mathematical functions for FDD. Additionally, \cite{Harinarayan2022} applied counterfactual explanation to deep learning-based models. Techniques in XAI, such as the Shapley value \citep{hart1989shapley}, have also been leveraged by \cite{bhakte2022explainable, bhakte2023alarm} to attribute the predictions of deep learning models to the contributions of each process variable.

However, ideally, one would like to have natural language-based systems to explain faults, root causes, and answer follow-up questions raised by users. Large Language Models (LLMs), such as GPT-4 \citep{OpenAI2023} and Llama-2 \citep{Touvron2023LlamaModels}, which excel at qualitative natural language tasks such as biology and bar exams.  However, these models struggle with more quantitative tasks such as answering math questions. Recently, OpenAI's o1 model \citep{openai_o1} enhances reasoning abilities, particularly in complex tasks like mathematics and coding, by employing extended ``thinking'' time before responding, which improves accuracy and reliability. These recent advances make LLM a promising tool for improving the interpretability of ML models in chemical engineering. 

To the best of our knowledge, an LLM-based agent for FDD has not yet been built. The closest work in this space is TalkToModel \citep{Slack2023ExplainingTalkToModel}, a recently developed chatbot that explains ML models by combining LLMs with external tools, such as counterfactual explanations. However, TalkToModel is restricted to small ML models and does not provide domain knowledge and context to the user. OptiChat\citep{chen2024diagnosing} is an LLM-powered natural language dialogue system for explaining optimization models. In chemical process design, LLM has been applied to flowsheet and control structure generation \citep{balhorn2024graph,hirtreiter2024toward}. LLMs have also been applied to explaining biomedical data, like in scChat \citep{lu2024scchat}, LLaVA-Med \citep{li2023llavamed}, and Med-PaLM \citep{tu2024towards}.

We use the Tennessee Eastman Process (TEP)\citep{downs1993plant} to demonstrate our method. TEP is a widely recognized benchmark for process control research and simulates an industrial plant with five major units and eight components. The process includes a reactor, condenser, compressor, separator, and stripper, producing liquid products through exothermic reactions. TEP features 41 measured and 12 manipulated variables, including feed rates, pressures, temperatures, and valve positions. Fault detection is critical in TEP, which has 21 preprogrammed faults, such as step changes in process variables or changes in reaction kinetics. Among these, 15 of the faults have known root causes. Faults are generated through a simulation program that operates under closed-loop conditions with a plant-wide control scheme. This setup allows for a realistic evaluation of fault detection methods, making TEP an ideal testbed for assessing various process monitoring techniques.

\section{Overview of \emph{FaultExplainer}'s Capabilities}\label{sec:overview}
\begin{figure}
    \centering
    \includegraphics[width=0.8\textwidth]{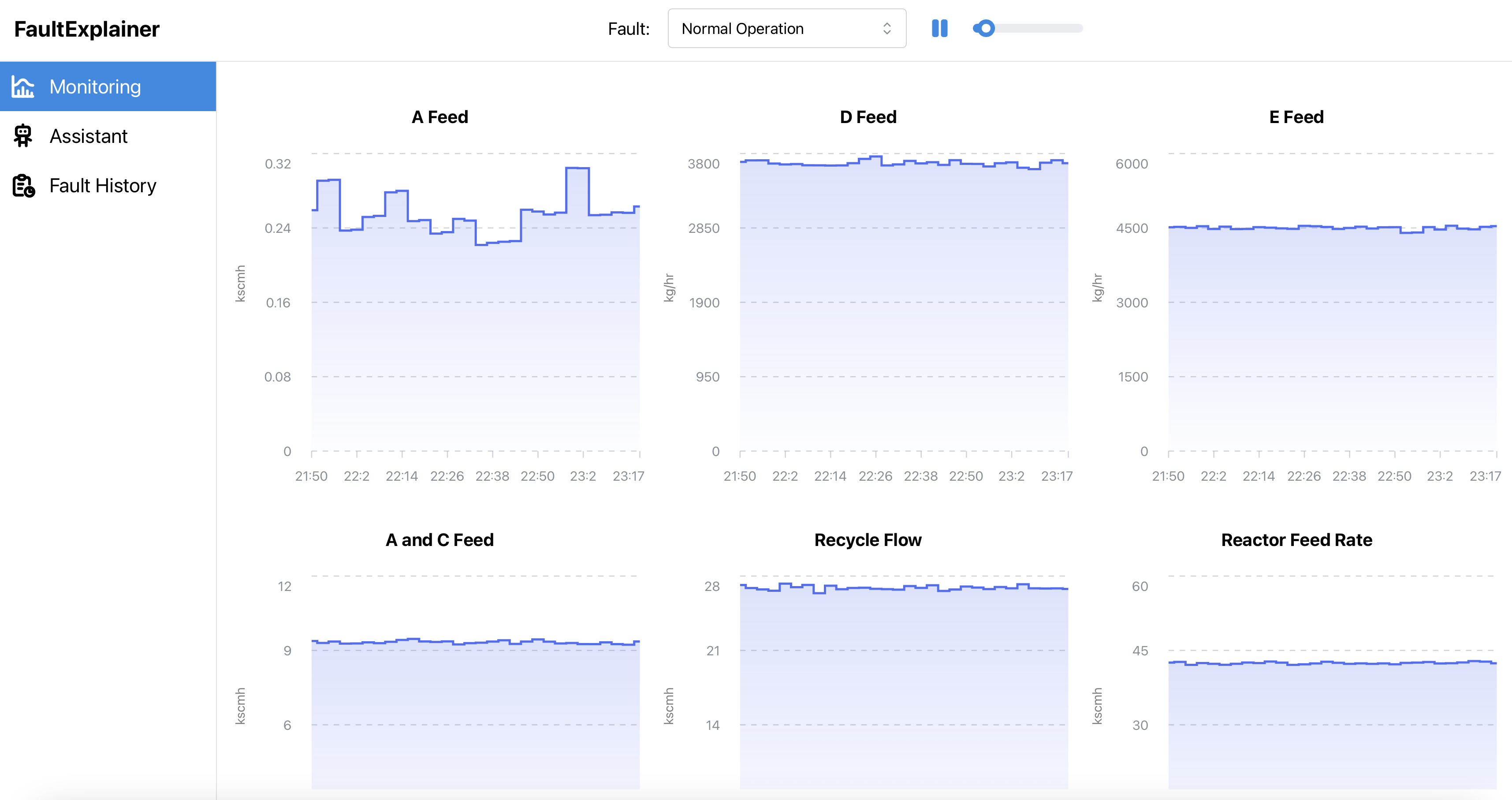}
    \includegraphics[width=0.8\textwidth]{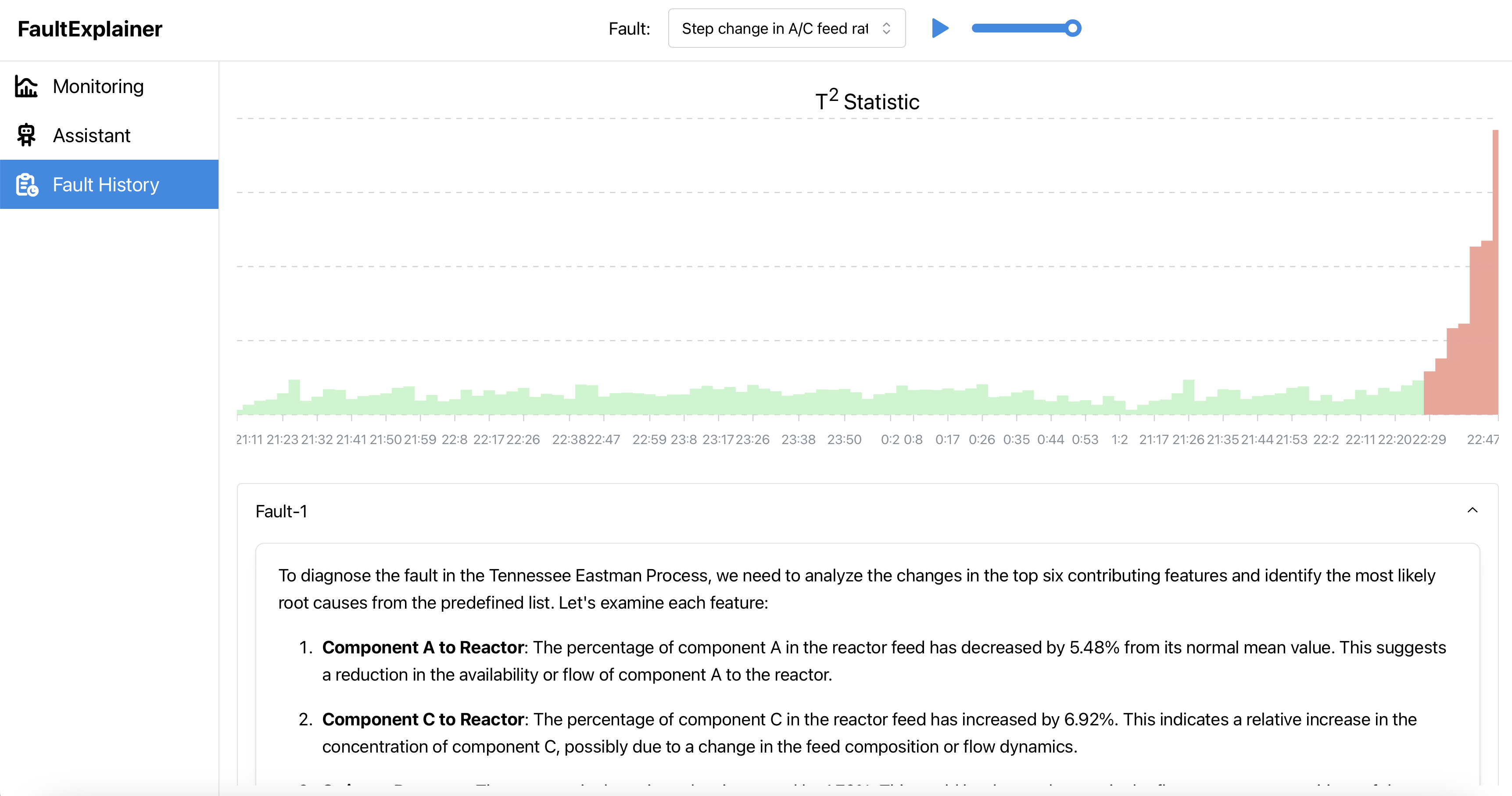}
       \includegraphics[width=0.8\textwidth]{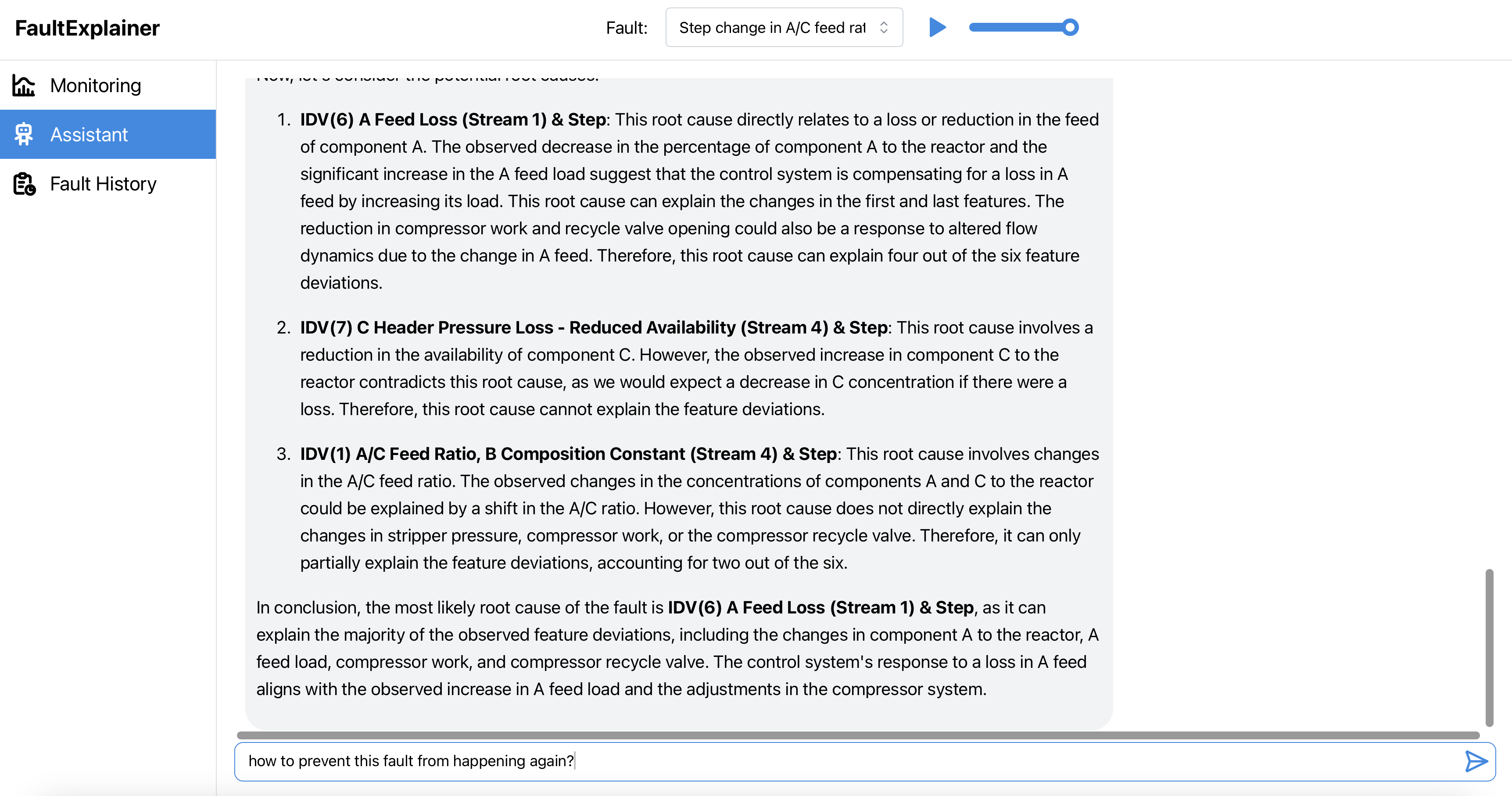}
    \caption{Interactive Web Interface for \emph{FaultExplainer}. From top to bottom are the process monitoring window, fault history tracker, and the interactive chat interface.}
    \label{fig:web_interface}
\end{figure}
To facilitate the practical application of our methodology and to make our findings accessible to a wider audience, we developed an interactive web interface for fault diagnosis and explanation within the Tennessee Eastman Process (TEP). 
Figure \ref{fig:web_interface} provides a visual overview of the capabilities of \emph{FaultExplainer} that can be accessed from the web interface.  It enables users to quickly identify and diagnose faults, and understand their underlying causes. This interface offers three key features corresponding to the three windows shown in Figure \ref{fig:web_interface}:
\begin{itemize}
    \item \textbf{Real-Time Sensor Data Visualization:} Users can monitor real-time data from all TEP sensors, enabling them to observe and track the behavior of the 41 measured variables and the 12 manipulated variables. The sensor data are generated from the TEP simulator. The user can introduce fault to the process by using the ``Fault'' dropdown button, e.g., changing the system from ``Normal Operation'' to a fault such as ``step change in A/C feed ratio''. All the faults described in \citep{downs1993plant} are available.
    \item \textbf{Fault History with $T^2$ Plot:} A dedicated ``Fault History'' page displays a real-time $T^2$ plot, along with a history of previously detected faults. This allows users to visualize the progression of faults over time and correlate them with changes in sensor data. The details of this method will be described in Section \ref{sec:methods}. After a fault is introduced, one can observe the change in $T^2$ statics in this window. When the $T_2$ statistics exceed the fault triggering threshold, the color in the $T^2$ plot will change from green to red. The historical fault reports contain the analysis of the major changes in sensor measurement, root cause deduction, and how the root cause propagates through the process.
    \item \textbf{Interactive Chat Interface:}  A chat interface provides a convenient way for users to inquire about specific faults or seek explanations for observed anomalies. Users can interact with the system using natural language, and the interface will generate grounded explanations based on the methodology presented in Section \ref{sec:methods}. The users can ask follow-up qualitative questions related to the process and the faults such as ``how to prevent the previous fault from occurring?''.
\end{itemize}

\section{Methods}\label{sec:methods}
\begin{figure}
    \centering
    \includegraphics[width=0.75\linewidth]{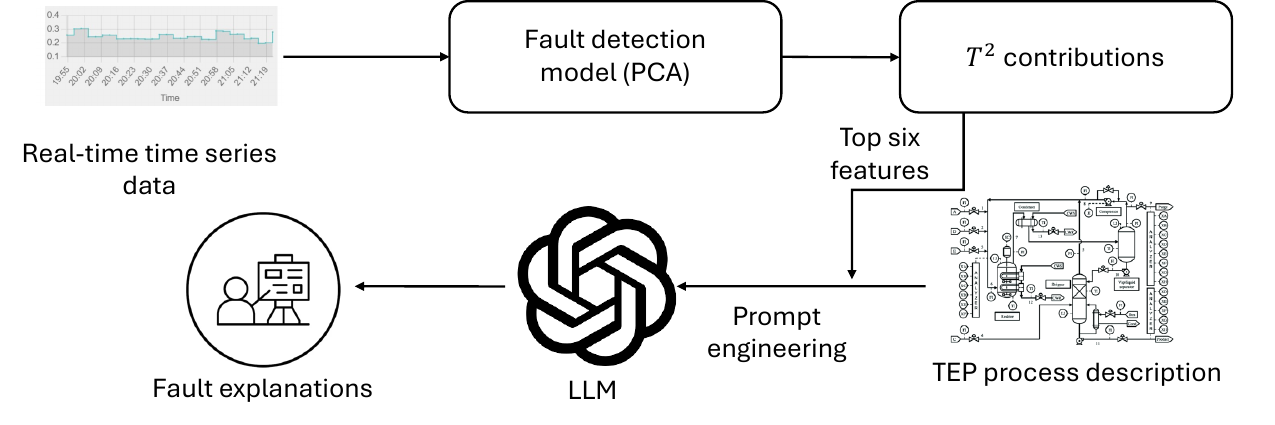}
    \caption{Overview of the Methods in \emph{FaultExplainer}}
    \label{fig:method}
\end{figure}

To alleviate the hallucinations of the LLMs, we propose an approach by adapting classical approaches for fault detection and feature importance analysis to enhance fault explanation in the Tennessee Eastman Process (TEP). The approach combines Principal Component Analysis (PCA) for identifying deviations in process data with $T^2$ contribution analysis to pinpoint relevant process variables. The outputs from these rigorous methods together with a detailed description of the TEP are used as input to the LLM. An overview of the proposed method is shown in Figure \ref{fig:method}. In what follows, we provide detailed explanations of the approach.

\subsection{Fault Detection and Feature Importance}
Faults in the Tennessee Eastman Process (TEP) are detected using a Principal Component Analysis (PCA)-based approach. PCA is a classical statistical learning technique used to reduce the dimensionality of a dataset while retaining most of the variability present in the data \citep{Chiang2000FaultSystems}. To make this paper self-contained, we provide a brief background of PCA. Mathematically, PCA involves several steps. First, the normal operating condition data is standardized by centering it (subtracting the mean of each variable) and scaling it (dividing by the standard deviation). The resulting $n$ scaled data points is represented by the matrix $X\in \mathcal{R}^{n \times m}$.  Then, the covariance matrix of the standardized data is computed. Eigendecomposition is performed on this covariance matrix to obtain the eigenvalues and eigenvectors. 
$$
S=\frac{1}{n-1} X^T X=V \Lambda V^T
$$

where the diagonal matrix $\Lambda \in \mathcal{R}^{m \times m}$ contains the non-negative real eigenvalues of decreasing magnitude ( $\lambda_1 \geq \lambda_2 \geq \cdots \geq \lambda_m \geq 0$ ). $V$ is the matrix that contains all the eigenvectors.

The eigenvectors corresponding to the largest eigenvalues form the principal components. In the PCA model, the original data matrix \(X\) can be decomposed into three matrices: \(X = TP^T + E\). Here, \(T=XP\) is the score matrix, which represents the projections of the original data onto the principal components. \(P\in\mathcal{R}^{m\times a}\) is the loading matrix, containing the loading vectors (principal components) associated only with the $a$ largest eigenvalues. \(E\) is the residual matrix, capturing the variations not explained by the principal components.

PCA can be used for fault identification through the $T^2$ statistic:

\begin{equation}
    T^2=x^T P \Lambda_a^{-1} P^T x
\end{equation}

where ${x}\in\mathcal{R}^{m}$ represents the scaled real-time sensor measurement, i.e., we assume the measurement has been standardized by the mean and standard deviation estimated from the normal operating condition data. $\Lambda_a$ represents the diagonal matrix that contains the $a$ largest eigenvalues. The $T^2$ statistic threshold is set to be,
\begin{equation}\label{eq:Tthreshold}
    T_\alpha^2=\frac{a(n-1)(n+1)}{n(n-a)} F_\alpha(a, n-a)
\end{equation}
where $\alpha$ is the probability threshold, which corresponds to the false alarm probability. $F$ represents the F-ratio distribution. When $T^2> T^{2}_{\alpha}$, a fault alarm will be triggered.

To discern the most influential variables contributing to these detected faults, we conducted a $T^2$ contribution analysis. This analysis quantifies the individual contributions of each variable to the overall $T^2$ statistic, allowing us to select the top contributing variables as candidates for further investigation. 

The contribution of $x_{j}$ can be calculated as 

\begin{equation}\label{eq:conttribution}
   CONT_{j}= \sum_{i=1}^{a}\frac{t_{i}}{\lambda_{i}}p_{j,i}x_{j}
\end{equation}
where $p_{j,i}$ is the $(j,i)$th element of the loading matrix $P$. $t_{i}$ is the $i$th element of the score $t=P^{T}x$, $\lambda_{i}$ is the $i$th largest eigenvalue. 

By decomposing the \(T^2\) statistic this way, we can identify how much each variable \( x_j \) contributes to the overall \(T^2\) value, which helps in pinpointing which variables are contributing the most to the detected fault. This detailed analysis assists in diagnosing the source of faults and taking targeted corrective actions.

\subsection{TEP Process Description and Prompt Engineering}\label{subsec:prompting}
To effectively use the LLM for fault explanation in the Tennessee Eastman Process (TEP), we first provide a detailed description of the process, including its unit operations, streams, and variables. This description clarifies the relationships among reactants, products, and byproducts, the role of each stream, and the monitored variables in the process. We explicitly distinguish between measured variables, which are recorded by sensors but cannot be directly manipulated, and manipulated variables, which can be actively adjusted by the control algorithm. This distinction allows the LLM to reason about process dynamics and the control system's responses to deviations. Additionally, the top six PCA contributing features (variables) calculated by Equation \eqref{eq:conttribution} and their deviations from their mean values during normal operating conditions are provided as inputs to guide the LLM in identifying significant deviations and their potential explanations.

Two different prompts are developed to support fault diagnosis.  The first prompt includes a predefined list of 15 root causes shown in the first column of Table \ref{tab:rootcauses}, retrieved from the original TEP paper \citep{downs1993plant}. This list acts as a constrained search space for the LLM, mimicking the scenario where the faults lie within historically encountered faults.  The second prompt is named \textit{Root Causes-Included Prompt}. The second prompt assumes no prior knowledge of the potential fault causes, allowing the LLM to generate hypotheses based solely on the feature deviations and process description, which is to mimic the scenario of encountering unseen faults. This prompt is called \textit{General Reasoning Prompt}. Both prompts are designed to elicit deterministic and consistent explanations by systematically analyzing the deviations, hypothesizing root causes, and connecting the identified faults to the observed feature changes. These carefully constructed prompts enable the LLM to leverage domain knowledge effectively while remaining grounded in the provided data and process details.

When the PCA identifies that a fault has occurred in the plant, the process description, the deviations of the top six contributing features from the normal operating condition, and one of the explanation prompts can be combined as inputs to the LLM. The LLM will generate a report describing what has been observed, the possible root causes of the faults, and how these root causes are propagated in the plant causing the variables to deviate. All the prompts are shown in the supplementary text.

\section{Results}\label{sec:results}
\subsection{Experiment Setup}
The time series data for the normal operating conditions and the 15 known faults of the TEP are obtained from \cite{DVN/6C3JR1_2017}. These data include 41 measured and 12 manipulated variables, with each time series containing 500 time steps. The normal operating time series is used to train the PCA model to extract principal components that capture 90\% of the data variance. The $T^2$ statistic threshold in Equation \ref{eq:Tthreshold} is set to $\alpha=0.01$, controlling the probability of a false fault report to $1\%$. \emph{FaultExplainer} triggers a fault only when six consecutive $T^2$ values exceed the threshold. Assuming independent observations, this corresponds to a false alarm rate of $0.01^6 = 1 \times 10^{-12}$. Once a fault is detected, we select the six features (variables) with the largest contribution values at the last time step. The values of these features are compared to their mean values during normal operation, and the value comparisons and percentage changes are incorporated into the prompts for the LLM.

We evaluate \emph{FaultExplainer} using two settings described in Section \ref{subsec:prompting}. In the \textit{Root Causes-Included Prompt} setting, the LLM is provided with a predefined list of 15 known root causes and must identify and explain the top three most likely causes based on the top six feature changes. In the \textit{General Reasoning Prompt} setting, the LLM is not restricted to predefined root causes and must independently infer and explain three potential root causes using its reasoning capabilities. The latter reflects real-world scenarios where faults may not have been previously encountered, testing the flexibility and robustness of the tool. 

We test OpenAI's GPT-4o \citep{openai_gpt4o} and o1-preview \citep{openai_o1} models. The GPT-4o model, a multimodal system, excels at general natural language tasks, while the o1-preview model incorporates improved reasoning capabilities. These experiments are designed to evaluate the models' accuracy, reasoning ability, and the quality of their explanations across both prompt settings. All outputs from both models for diagnosing the 15 faults under the two prompt settings are provided in the supplementary material.

\subsection{Results of the Root Causes-Included Prompt}
We test the \textit{Root Causes-Included Prompt} over the 15 faults. Both LLMs are asked to give the top three possible fault classifications.
\subsubsection{Quantitative evaluation of the classification accuracy}

Table \ref{tab:rootcauses} summarizes the fault classification results of the GPT-4o and o1-preview models when prompted with the 15 predefined root causes. The numbers in bold indicate correct classifications. For certain faults, such as Faults 1 and 8, the contributing features alone cannot distinguish between random variations and step changes in the same features. These faults are considered aliases and are treated as correct as long as the model classifies the fault or its alias correctly. Faults that cannot be detected by PCA are marked with backslashes in the table.

Among the 11 faults that can be identified using PCA, GPT-4o correctly classifies 7 faults, while o1-preview correctly classifies 9 faults. This aligns with o1-preview's improved reasoning capabilities, allowing it to better utilize the provided descriptions and contributing features. Neither model is able to correctly diagnose Faults 5 (Condenser Cooling Water Inlet Temperature \& Step) and 10 (C Feed Temperature (Stream 4) \& Random Variation). This failure is mainly because the LLMs find alternative ways to explain the deviations of the top six features, which will be discussed in detail in the qualitative evaluation of the fault reports. 

\begin{table}[htbp]
\scriptsize
  \centering
  \caption{Summary of GPT-4o and o1-preview's top three fault classification when prompted with the 15 root faults. The correct classifications are denoted in bold. The backslashes mean the corresponding faults cannot be detected by the PCA.}\label{tab:rootcauses}
    \begin{tabular}{l|l|l}
    \toprule
    Faults & GPT-4o & o1-preview \\
    \midrule
    Fault 1 A/C Feed Ratio, B Composition Constant (Stream 4) \& Step (alias 8) & 6,7,4 & 6,\textbf{1},7 \\
    \midrule
    Fault 2 B Composition, A/C Ratio Constant (Stream 4) \& Step  (alias 8) & 6,1,\textbf{8} & \textbf{2},6,7 \\
    \midrule
    Fault 3 D Feed Temperature (Stream 2) \& Step (alias 9) & \textbackslash{} & \textbackslash{} \\
    \midrule
    Fault 4 Reactor Cooling Water Inlet Temperature \& Step (alias 11,14) & \textbackslash{} & \textbackslash{} \\
    \midrule
    Fault 5 Condenser Cooling Water Inlet Temperature \& Step (alias 12,15) & 3,4,6 & 4,7,1 \\
    \midrule
    Fault 6 A Feed Loss (Stream 1) \& Step & \textbf{6},7,4 & \textbf{6},7,1 \\
    \midrule
    Fault 7 C Header Pressure Loss - Reduced Availability (Stream 4) \& Step & \textbf{7},6,8 & \textbf{7},6,1 \\
    \midrule
    Fault 8 A, B, C Feed Composition (Stream 4) \& Random Variation (alias 1,2) & 6,7,4 & 6,7,\textbf{1} \\
    \midrule
    Fault 9 D Feed Temperature (Stream 2) \& Random Variation (alias 3) & \textbackslash{} & \textbackslash{} \\
    \midrule
    Fault 10 C Feed Temperature (Stream 4) \& Random Variation & 6,1,4 & 13,1,7 \\
    \midrule
    Fault 11 Reactor Cooling Water Inlet Temperature \& Random Variation (alias 4,14) & \textbf{4},5,6 & \textbf{4},3,\textbf{14} \\
    \midrule
    Fault 12 Condenser Cooling Water Inlet Temperature \& Random Variation (alias 5, 15) & 4,\textbf{5},6 & 2,13,\textbf{5} \\
    \midrule
    Fault 13 Reaction Kinetics \& Slow Drift & 4,6,\textbf{13} & 3,\textbf{13},6 \\
    \midrule
    Fault 14 Reactor Cooling Water Valve \& Sticking (alias 4, 11) & \textbf{4},6,5 & \textbf{4},\textbf{14},7 \\
    \midrule
    Fault 15 Condenser Cooling Water Valve \& Sticking (alias 5, 15) & \textbackslash{} & \textbackslash{} \\
    \bottomrule
    \end{tabular}%
  \label{tab:addlabel}%
\end{table}%
\subsubsection{Qualitative evaluation of the explanations}\label{sec:qualitativerootcauses}
In this subsection, we pick a fault where both models find the correct root cause and one fault where neither model finds the correct root cause to demonstrate the advantages and limitations of the two models. The rest of the fault reports follow a similar pattern and can be found in the supplementary material.

The results for Fault 7, where both GPT-4o and o1-preview identify the correct root cause (C Header Pressure Loss - Reduced Availability in Stream 4), shown in Figure \ref{fig:fault7sucessrootcause}, demonstrate the plausibility and coherence of the explanations provided by both models. Both models accurately link the observed decreases in the reactor, product separator, and stripper pressures, as well as the reduced A and C feed flow rate, to the loss in header pressure. They also correctly attribute the control system's compensatory behavior—namely, the increase in the A and C feed load—to the attempt to counteract the pressure loss. Furthermore, the explanations addressed the decrease in compressor work as a consequence of reduced flow rates and vapor generation.
\begin{figure}[h!]
    \centering
    \resizebox{\textwidth}{!}{
    \begin{tikzpicture}[
        query/.style = {rectangle, draw=blue!70, fill=blue!15, rounded corners, text width=13cm, align=center, font=\footnotesize, node distance=1.5cm},
        action/.style = {rectangle, draw=green!70, fill=green!15, rounded corners, text width=13cm, align=center, font=\footnotesize, node distance=1.5cm},
                o1/.style = {rectangle, draw=green!70, fill=green!15, rounded corners, text width=13cm, align=center, font=\footnotesize, node distance=1.5cm},
    ]
     \node[query](c) {\textbf{Top six feature changes}\\
   We analyze the percentage changes in the top six contributing features during the fault compared to normal operation. The \textbf{Stripper Pressure} decreased by $3.00\%$, the \textbf{A and C Feed} flow rate decreased by $8.90\%$, the \textbf{Reactor Pressure} decreased by $3.49\%$, and the \textbf{Product Separator Pressure} decreased by $3.69\%$. The manipulated variable \textbf{A and C Feed Load} increased by $13.12\%$, and the \textbf{Compressor Work} decreased by $3.58\%$.
    };

    \node[action, below=0.1cm of c](a) {\textbf{GPT-4o explanation}\\
    \textbf{IDV(7) C Header Pressure Loss - Reduced Availability (Stream 4) \& Step:} This root cause can explain the decrease in the A and C feed, as a pressure loss in the C header would directly reduce the availability of C, leading to a lower feed rate. The control system might respond by increasing the A and C feed load to compensate, explaining the observed increase in the feed load. The reduced availability of C could also lead to lower reaction rates, causing a decrease in reactor pressure and subsequently affecting the pressures in the stripper and product separator. The reduced flow rates could also lead to decreased compressor work.
    };
        \node[o1, below=0.1cm of a](b) {\textbf{o1-preview explanation}\\
         \textbf{IDV(7): C Header Pressure Loss - Reduced Availability (Stream 4) \& Step} A sudden loss in the C header pressure would reduce the availability of component C in Stream 4. This reduction directly causes the \textbf{A and C Feed} flow rate to decrease by $8.90\%$, as observed. In response, the control system increases the \textbf{A and C Feed Load} by $13.12\%$ in an attempt to maintain the flow rate, but due to the pressure loss, the flow cannot be restored to normal levels. The decreased feed rate leads to lower pressures in the \textbf{Reactor}, \textbf{Product Separator}, and \textbf{Stripper}, explaining the $3\%$ to $3.7\%$ decreases in pressures. The reduction in overall flow rates causes a decrease in vapor generation and thus a decrease in \textbf{Compressor Work} by $3.58\%$. Therefore, this root cause explains all six of the top feature deviations.
    };

    \end{tikzpicture}
    }
    \caption{An illustrative example where both models are able to identify the correct root causes. The top six feature changes and the explanation of GPT-4o and o1-preview to the top six feature changes of Fault 7 (C Header Pressure Loss - Reduced Availability (Stream 4) \& Step) when the 15 root causes are provided.}\label{fig:fault7sucessrootcause}
\end{figure}

Both models succeed in providing consistent and plausible interpretations of the observed feature changes, effectively connecting the root cause to the system's response and the measured deviations. \begin{figure}[h!]
    \centering
    \resizebox{\textwidth}{!}{
    \begin{tikzpicture}[
        query/.style = {rectangle, draw=blue!70, fill=blue!15, rounded corners, text width=13cm, align=center, font=\footnotesize, node distance=1.5cm},
        action/.style = {rectangle, draw=green!70, fill=green!15, rounded corners, text width=13cm, align=center, font=\footnotesize, node distance=1.5cm},
                o1/.style = {rectangle, draw=green!70, fill=green!15, rounded corners, text width=13cm, align=center, font=\footnotesize, node distance=1.5cm},
    ]
     \node[query](c) {\textbf{Top six feature changes}\\
The \textbf{Stripper Temperature} increased by $2.45\%$, the \textbf{A Feed Load} increased by $43.90\%$, and the \textbf{A Feed Rate} increased by $43.63\%$. Meanwhile, the \textbf{E Feed Rate} decreased by $2.38\%$. The \textbf{Stripper Steam Valve} position increased by $14.59\%$, and the \textbf{Stripper Steam Flow} increased by $9.93\%$.

    };

    \node[action, below=0.1cm of c](a) {\textbf{GPT-4o explanation}\\
 \textbf{IDV(6) A Feed Loss (Stream 1) \& Step}: This root cause could explain the significant increase in the A feed load and A Feed, as the control system might compensate for a perceived loss by increasing the feed rate. \textit{The increase in A feed could lead to changes in the reactor's operation, affecting the stripper temperature and steam flow as the system attempts to maintain product quality.} However, this root cause does not directly explain the decrease in E Feed. Therefore, this root cause can explain 5 out of the 6 feature deviations.
    };
        \node[o1, below=0.1cm of a](b) {\textbf{o1-preview explanation}\\
        The most plausible root cause is \textbf{IDV(13): Reaction Kinetics \& Slow Drift}. A gradual deterioration of reaction kinetics, possibly from catalyst deactivation or fouling, would reduce the conversion rates of reactants to products. To maintain the desired production levels, the control system responds by increasing the \textbf{A feed load} and \textbf{A Feed}, supplying more reactant A to the reactor. This explains the significant increases in these features. The slight decrease in \textbf{E Feed} may result from an accumulation of unreacted E in the reactor due to slower reaction rates, prompting the control system to reduce its feed rate.

The reduced reaction efficiency leads to higher concentrations of unreacted reactants in the reactor effluent. Consequently, the control system increases the \textbf{Stripper steam valve} opening and the \textbf{Stripper Steam Flow} to enhance the removal of residual reactants from the product stream. This accounts for the observed increases in these features. The increased steam flow raises the \textbf{Stripper Temperature}, aligning with the observed temperature rise. All six of the top features can be explained by \textbf{IDV(13)}.};

    \end{tikzpicture}
    }
    \caption{An illustrative example where neither model is able to identify the correct root cause. The top six feature changes and the explanation of GPT-4o and o1-preview to the top six feature changes of Fault 10 C Feed Temperature (Stream 4) \& Random Variation when the 15 root causes are provided. The hallucinated explanations are shown in italics.}\label{fig:fault10failturerootcause}
\end{figure}
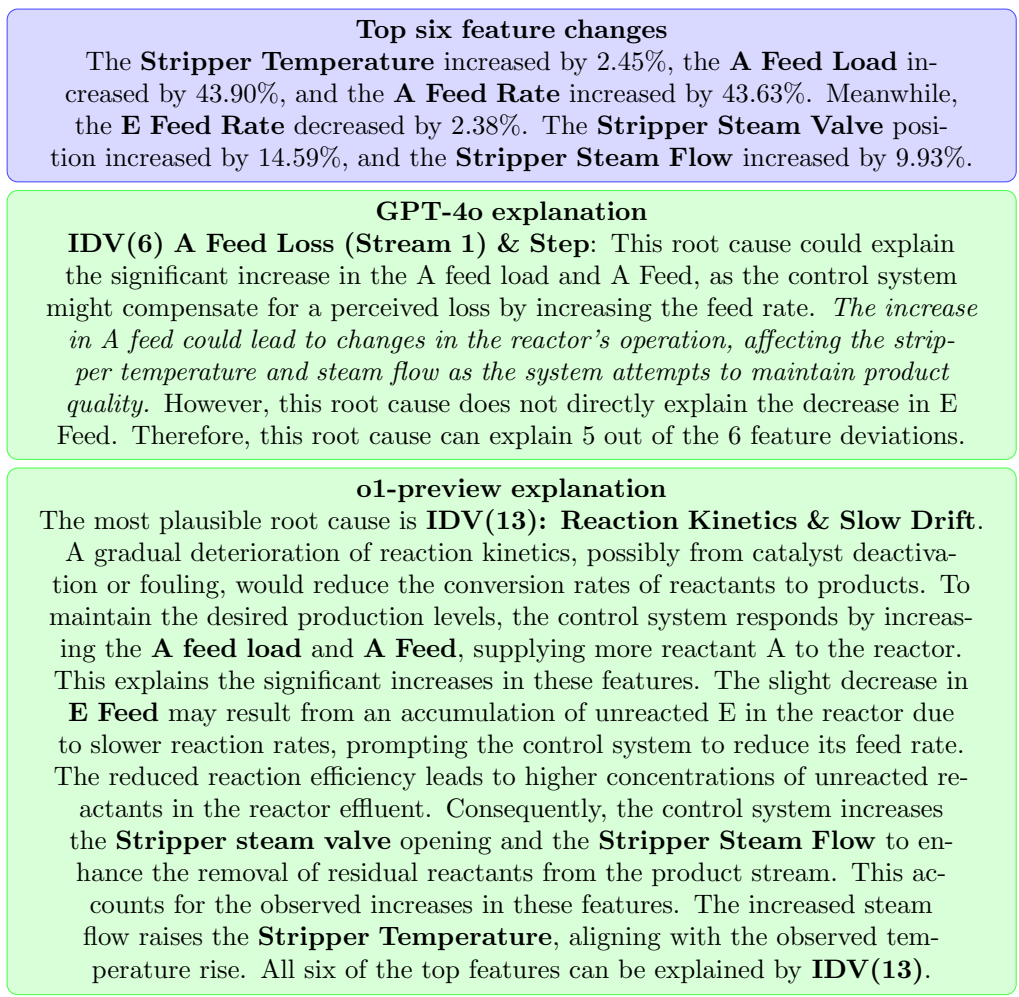
The explanations also align well with the expected behavior of the TEP control system, which adjusts manipulated variables to maintain operational stability. Notably, o1-preview offered a more detailed sequence of events, explicitly tying each feature deviation to the control system's response and providing a clear causal pathway for fault propagation. This demonstrates o1-preview's advantage in reasoning and generating deterministic, system-oriented explanations, although both models were successful in this case.

Figure \ref{fig:fault10failturerootcause} shows the failure of both GPT-4o and o1-preview to identify the correct root cause (Fault 10: C Feed Temperature Random Variation). It highlights important limitations in both the LLMs' reasoning and the feature selection process. GPT-4o's explanation lacked quantitative rigor and causal clarity, failing to explicitly connect the observed feature changes to the hypothesized root cause. Instead, it provides a vague connection between the increase in A feed load and A feed rate to the changes in stripper steam flow and temperature, which appears to be a hallucination rather than a reasoned deduction. This hallucinated explanation is highlighted in italics in Figure \ref{fig:fault10failturerootcause}. This lack of precision resulted in an incomplete and unconvincing explanation that do not account for all observed feature changes.

On the other hand, o1-preview produced a plausible and systematic explanation based on chemical engineering knowledge, attributing the deviations to reaction kinetics and slow drift (IDV(13)). However, despite its coherence, this explanation was incorrect because the identified features—such as the stripper steam flow and A Feed—are not directly related to the actual root cause (C Feed Temperature Random Variation). The model's reasoning was limited by the feature set provided by PCA, which failed to capture variables directly linked to the root cause. Consequently, o1-preview attempted to construct an alternative causal pathway to explain the deviations, ultimately leading to a misclassification.
This failure underscores the dual challenges in fault diagnosis: (1) the need for the LLM to reason effectively even when feature relationships to the root cause are indirect, and (2) the limitations of PCA in identifying features that are strongly tied to the true root cause. Addressing these issues requires improving both feature selection methodologies and LLM reasoning capabilities to ensure robust and accurate fault explanations.
\subsection{Results of the General Reasoning Prompt}
Similarly, the \textit{General Reasoning Prompt} is used to generate the top three root causes with explanations.
\subsubsection{Quantitative evaluation of the classification accuracy}
\begin{table}[htbp]
\scriptsize
  \centering
  \caption{A summary of the top three root causes generated by GPT-4o and o1-preview using the general reasoning prompt. The root causes related to the ground truth of each fault are highlighted in bold.}
    \begin{tabular}{c|c|c}
    \toprule
    Faults & \multicolumn{1}{l|}{GPT-4o} & \multicolumn{1}{l}{o1-preview} \\
    \midrule
    1     & \multicolumn{1}{p{21.415em}|}{\textbf{Malfunction in the Feed System A}\newline{}Control System Malfunction \newline{}\textbf{Process Stream Disturbance}} & \multicolumn{1}{p{25.335em}}{\textbf{Increased Component C in Feed Stream 4}\newline{}Malfunction in the Recycle Stream\newline{}\textbf{Partial Blockage in Component A Feed Line} } \\
    \midrule
    2     & \multicolumn{1}{p{21.415em}|}{Increased Reactor Pressure or Flow Rate\newline{}Malfunction in the Vapor-Liquid Separator\newline{}Change in Feed Composition or Quality} & \multicolumn{1}{p{25.335em}}{\textbf{Increased Ingress of Inert B into the System}\newline{}Condenser Inefficiency Leading to Reduced Condensation\newline{}Compressor Malfunction Leading to Reduced Recycle Flow} \\
    \midrule
    3     & \textbackslash{} & \textbackslash{} \\
    \midrule
    4     & \textbackslash{} & \textbackslash{} \\
    \midrule
    5     & \multicolumn{1}{p{21.415em}|}{Increased Reaction Temperature\newline{}\textbf{Malfunction in Cooling System}\newline{}Feed Composition Change} & \multicolumn{1}{p{25.335em}}{Decrease in E Feed Supply\newline{}\textbf{Cooling Inefficiency in the Separator Condenser}\newline{}Decrease in Component C Supply to Reactor} \\
    \midrule
    6     & \multicolumn{1}{p{21.415em}|}{\textbf{Malfunction in A Feed Control System}\newline{}Mechanical Issue in Compressor\newline{}Supply Issue with Reactant E} & \multicolumn{1}{p{25.335em}}{Compressor Recycle Valve Malfunction\newline{}Compressor Underperformance Due to Mechanical Fault\newline{}\textbf{Partial Blockage in A Feed Line} } \\
    \midrule
    7     & \multicolumn{1}{p{21.415em}|}{\textbf{Feed Supply Disruption}\newline{}Reactor Catalyst Deactivation\newline{}Cooling System Malfunction} & \multicolumn{1}{p{25.335em}}{\textbf{Partial Blockage in A and C Feed Line}\newline{}\textbf{Malfunctioning A and C Feed Control Valve}\newline{}Leak in the Reactor System} \\
    \midrule
    8     & \multicolumn{1}{p{21.415em}|}{\textbf{Malfunction in the A Feed System}\newline{}\textbf{Control System Malfunction}\newline{}Issue with the Recycle Compressor} & \multicolumn{1}{p{25.335em}}{\textbf{Decrease in Component C Feed}\newline{}Catalyst Deactivation in Reactor\newline{}Instrumentation Error in Component C Measurements} \\
    \midrule
    9     & \textbackslash{} & \textbackslash{} \\
    \midrule
    10    & \multicolumn{1}{p{21.415em}|}{Increased A Feed Rate\newline{}Control System Malfunction\newline{}Stripper Performance Issue} & \multicolumn{1}{p{25.335em}}{Decrease in E Feed Rate Due to Supply Limitation\newline{}Increase in A Feed Setpoint \newline{}Operational Issue in the Stripper} \\
    \midrule
    11    & \multicolumn{1}{p{21.415em}|}{\textbf{Increased Reaction Rates Due to Temperature Rise}\newline{}Inefficient Cooling in the Separator\newline{}Feed Composition Imbalance} & \multicolumn{1}{p{25.335em}}{Increase in D Feed Rate\newline{}\textbf{Blockage in Reactor Cooling Water Valve}\newline{}Increase in D Feed Temperature} \\
    \midrule
    12    & \multicolumn{1}{p{21.415em}|}{\textbf{Inefficient Heat Removal in the Separator}\newline{}Change in Feed Composition\newline{}Reactor Operating Conditions} & \multicolumn{1}{p{25.335em}}{Increase in Inert Component B in Feed Stream 4\newline{}Decrease in A or C Concentration in Feed Stream 4\newline{}Decreased Reactor Catalyst Activity} \\
    \midrule
    13    & \multicolumn{1}{p{21.415em}|}{Feed Composition Change\newline{}Reactor Malfunction\newline{}Control System Malfunction} & \multicolumn{1}{p{25.335em}}{Decrease in D Feed Rate (Stream 2)\newline{}Blockage or Restriction in D Feed Line\newline{}Malfunctioning Controller Decreasing D Feed Load} \\
    \midrule
    14    & \multicolumn{1}{p{21.415em}|}{\textbf{Cooling System Malfunction}\newline{}Feed Composition Change\newline{}Reaction Kinetics Shift} & \multicolumn{1}{p{25.335em}}{\textbf{Decreased Reactor Cooling Efficiency} \newline{}Increase in Inert Component B Feed to Reactor\newline{}Decrease in Component C Feed to Reactor} \\
    \midrule
    15    & \textbackslash{} & \textbackslash{} \\
    \bottomrule
    \end{tabular}%
  \label{tab:generalresonsingresults}%
\end{table}%
The results of \textit{General Reasoning Prompt}, summarized in Table \ref{tab:generalresonsingresults}, highlight the performance of GPT-4o and o1-preview when not restricted to the 15 predefined root causes. The root causes generated by the LLMs that are related to the ground truth of the faults are highlighted in bold in Table \ref{tab:generalresonsingresults}. Both models identified the correct or related root cause in 8 out of the 11 faults identified by the PCA. This demonstrates their capacity to generalize reasoning beyond fixed fault lists, which is critical in practical scenarios where previously unseen faults may arise. However, neither model is able to identify the correct root causes for Faults 10 and 13, illustrating a key limitation. In these cases, the inability to correlate the observed feature deviations with the ground truth highlights the challenge of linking indirect feature changes to the actual fault mechanism. We will discuss these limitations in detail in our qualitative evaluation.

\subsubsection{Qualitative evaluation of the explanations}
Similar to Section \ref{sec:qualitativerootcauses}, we choose two faults to demonstrate the advantages and limitations of the two models. The rest of the fault reports follow a similar pattern and can be found in the supplementary material.
\begin{figure}[h!]
    \centering
    \resizebox{\textwidth}{!}{
    \begin{tikzpicture}[
        query/.style = {rectangle, draw=blue!70, fill=blue!15, rounded corners, text width=13cm, align=center, font=\footnotesize, node distance=1.5cm},
        action/.style = {rectangle, draw=green!70, fill=green!15, rounded corners, text width=13cm, align=center, font=\footnotesize, node distance=1.5cm},
                o1/.style = {rectangle, draw=green!70, fill=green!15, rounded corners, text width=13cm, align=center, font=\footnotesize, node distance=1.5cm},
    ]
     \node[query](c) {\textbf{Top six feature changes}\\
The \textbf{Component B in Purge} increased by $5.63\%$, the \textbf{Purge Rate} by $20.96\%$, and the \textbf{Purge Valve} by $20.86\%$. Additionally, \textbf{Component B to Reactor} increased by $4.17\%$, the \textbf{E Feed Load} by $1.67\%$, and the \textbf{A Feed Load} showed the most significant increase at $30.74\%$.

    };

    \node[action, below=0.1cm of c](a) {\textbf{GPT-4o explanation}\\
    \textbf{Root Cause 1: Increased Reactor Pressure or Flow Rate:} An increase in reactor pressure or flow rate could lead to higher purge rates and changes in the composition of the purge stream. The control system might respond by increasing the purge valve opening to manage the pressure, leading to the observed increase in purge rate and component B in the purge. \textit{The increased flow rate could also result in more component B being carried over to the reactor, explaining the increase in component B to the reactor. The increase in A and E feed loads could be a response to maintain reaction rates and product output under the new conditions.} This root cause can explain all six feature deviations.
    };
        \node[o1, below=0.1cm of a](b) {\textbf{o1-preview explanation}\\
        \textbf{Root Cause 1: Increased Ingress of Inert \textit{B} into the System}

An unexpected increase in the influx of inert Component \textit{B} into the process could explain the observed feature deviations. This may result from contamination in one of the feed streams, particularly Stream 4, which includes a mixture of \textit{A}, \textit{B}, and \textit{C}. If the concentration of \textit{B} in this stream increases, more inert \textit{B} enters the reactor feed, leading to an elevated mole percentage of Component \textit{B} to the Reactor (Feature 4), as observed (+4.17\%). The accumulation of inert \textit{B} in the process causes higher concentrations of \textit{B} in the recycle loop. To prevent excessive buildup of inert gas, the control system responds by increasing the Purge Valve opening (Feature 3, +20.86\%) to remove the excess inert. This action increases the Purge Rate (Feature 2, +20.96\%) and results in higher concentrations of Component \textit{B} in the Purge stream (Feature 1, +5.63\%).

The increased presence of inert \textit{B} in the reactor feed dilutes the concentrations of reactive species, potentially reducing the overall reaction rates. The control system compensates for this by increasing the feed rates of reactants \textit{A} and \textit{E}. This is evidenced by the significant increase in \textit{A} Feed Load (Feature 6, +30.74\%) and the smaller increase in \textit{E} Feed Load (Feature 5, +1.67\%). Since \textit{A} participates in all reactions and is more critical, a larger increase in its feed rate is justified. Therefore, this root cause explains all six observed feature deviations.

\textbf{Features explained:} 6/6.
    };

    \end{tikzpicture}
    }
    \caption{The top six feature changes and the explanation of GPT-4o and o1-preview to the top six feature changes of Fault 2 (B Composition, A/C Ratio Constant (Stream 4) \& Step) when the 15 root causes are not provided. The hallucinated explanations are highlighted in italics.}\label{fig:fault2generalreaonsing}
\end{figure}

Figure~\ref{fig:fault2generalreaonsing} shows how GPT-4o and o1-preview explain Fault 2 when the 15 predefined root causes are not provided. GPT-4o’s explanation lacks specificity and tends toward vague reasoning. While it proposes a scenario involving an increased reactor pressure or flow rate, it does not identify the affected components or streams precisely. Furthermore, it fails to establish a clear causal relationship between the proposed root cause and the observed feature changes. For example, it claims that an increase in reactor flow rate \emph{results in} more component B being flowed to the reactor. This is problematic because component B itself represents the reactor’s inflow, making the proposed causal linkage nonsensical. Furthermore, it claims that ``The increase in A and E feed loads could be a response to maintain reaction rates and product output under the new conditions.'' However, it does not explain why the new conditions make the A and E feed loads increase. The response suggests that GPT-4o may be compensating for insufficient reasoning with overly general statements.

In contrast to GPT-4o, o1-preview provides a thorough, causally grounded explanation that aligns closely with both the observed data and fundamental chemical engineering principles. As shown in the figure, o1-preview pinpoints an increased ingress of inert Component B into the system.
The increase in B causes the rise of B in the recycle loop, prompting the control system to mitigate the accumulation of B by increasing the Purge Valve opening and consequently the Purge Rate and the concentration of B in the purge.  The explanation also addresses the reactive species balance within the reactor. With inert B diluting the reactor, the control system compensates by increasing the feed loads of reactants A and E. o1-preview is able to explain all the six feature deviations from the normal operating conditions. Overall, the o1-preview’s explanation is more granular, causal, and coherent compared with GPT-4o.

\begin{figure}[h!]
    \centering
    \resizebox{\textwidth}{!}{
    \begin{tikzpicture}[
        query/.style = {rectangle, draw=blue!70, fill=blue!15, rounded corners, text width=13cm, align=center, font=\footnotesize, node distance=1.5cm},
        action/.style = {rectangle, draw=green!70, fill=green!15, rounded corners, text width=13cm, align=center, font=\footnotesize, node distance=1.5cm},
                o1/.style = {rectangle, draw=green!70, fill=green!15, rounded corners, text width=13cm, align=center, font=\footnotesize, node distance=1.5cm},
    ]
     \node[query](c) {\textbf{Top six feature changes}\\
   The \textbf{Stripper Pressure} increased by $1.58\%$, the \textbf{Reactor Pressure} increased by $1.84\%$, and the \textbf{Product Separator Pressure} increased by $1.92\%$. In contrast, the \textbf{Component D in Purge} decreased by $17.85\%$, the \textbf{Component D to Reactor} decreased by $3.31\%$, and the \textbf{Component F in Purge} decreased by $2.17\%$.

    };

    \node[action, below=0.1cm of c](a) {\textbf{GPT-4o explanation}\\
 \textbf{Root Cause 1: Feed Composition Change} \\
    A change in the feed composition, particularly a reduction in component D, could explain several of the observed deviations. The decrease in component D to the reactor and in the purge stream suggests that less D is available for reaction, potentially due to a feed supply issue or a change in the feed stream composition. \textit{This could lead to increased reactor pressure as the reaction rates are affected, causing a buildup of unreacted gases. The increased pressure in the reactor could propagate downstream, affecting the product separator and stripper pressures.} This root cause can explain the deviations in the reactor pressure, product separator pressure, component D in purge, and component D to reactor, accounting for 4 out of the 6 features.
    };
        \node[o1, below=0.1cm of a](b) {\textbf{o1-preview explanation}\\
        \textbf{Root Cause 1: Decrease in D Feed Rate (Stream 2)} A decrease in the D feed rate through Stream 2 directly leads to a reduction in the amount of Component D entering the reactor, explaining the 3.31\% decrease in Component D to Reactor. With less D fed into the reactor, there is less unreacted D available to exit via the purge stream, accounting for the significant 17.85\% decrease in Component D in Purge. The shortage of D affects Reactions 1 and 4, which consume D, resulting in decreased reaction rates. This decrease leads to an accumulation of unreacted gaseous reactants, such as A and C, increasing the reactor pressure by 1.84\%. The elevated reactor pressure propagates downstream, causing increases in the product separator pressure by 1.92\% and stripper pressure by 1.58\%. Additionally, the reduced availability of D diminishes the production of byproduct F through Reaction 4 (3D~$\rightarrow$~2F), explaining the slight 2.17\% decrease in Component F in Purge. Therefore, this root cause explains all six observed feature deviations.
        
    };

    \end{tikzpicture}
    }
    \caption{An illustrative example where neither model is able to identify the correct root cause. The top six feature changes and the explanation of GPT-4o and o1-preview to the top six feature changes of Fault 13 Reaction Kinetics \& Slow Drift when the 15 root causes are not provided. The hallucinated explanations are
highlighted in italics.}\label{fig:generalreasoningfailure}
\end{figure}
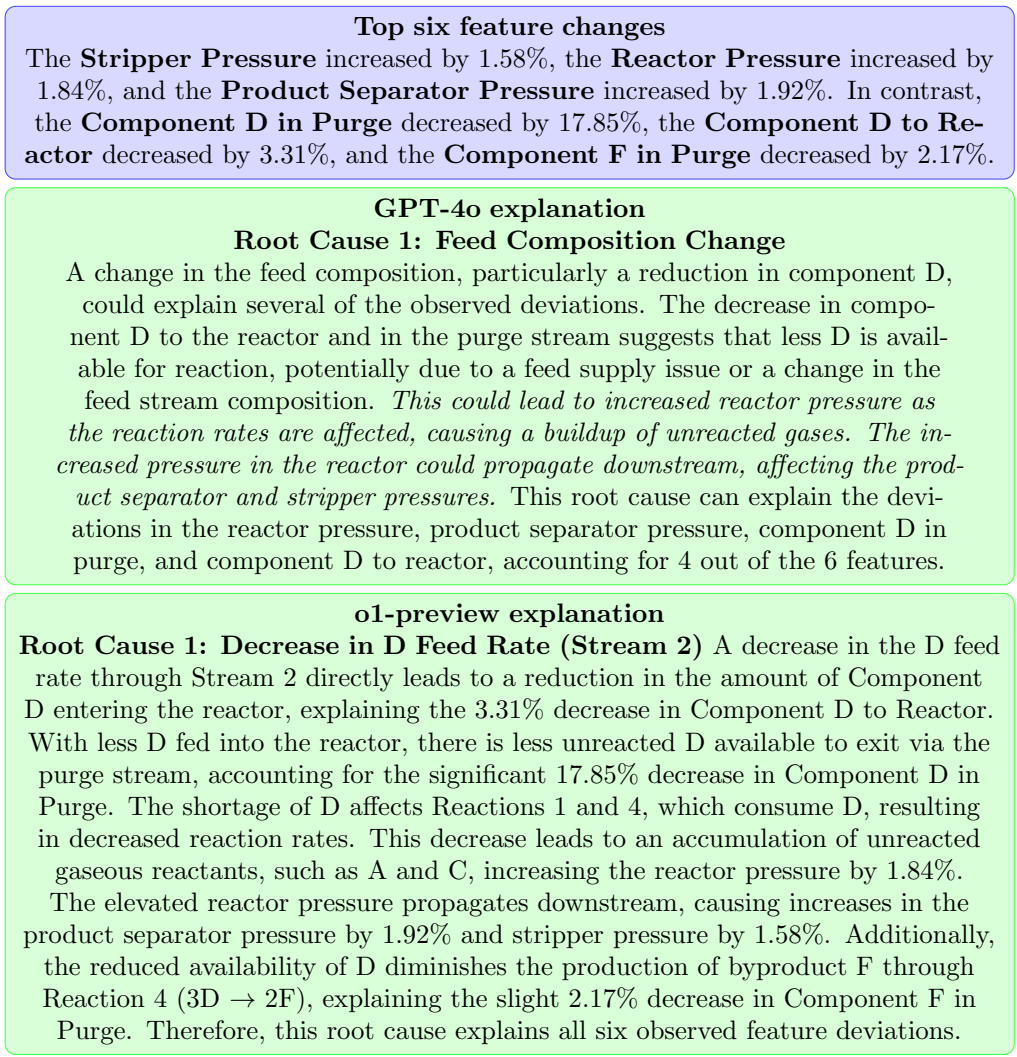
Figure \ref{fig:generalreasoningfailure} showcases a scenario where neither GPT-4o nor o1-preview is able to identify the correct root cause for Fault 13 (Reaction Kinetics \& Slow Drift). The top six feature changes, primarily related to pressure deviations and changes in Component D and F concentrations, are misattributed to alternative root causes by both models.

GPT-4o attributes the deviations to a D feed increase, providing an explanation for four out of the six feature changes. However, its explanation lacks justification and shows signs of hallucination. The hallucinated sentences are highlighted in italics. For example, it claims that the decrease in feed D to the reactor ``could lead to increased reactor pressure as the reaction rates are affected'' but fails to justify why the reactor pressure gets increased due to the decrease in feed D. It further claims that ``The increased pressure in the reactor could propagate downstream, affecting the product separator and stripper pressures'' but does not reason explicitly how the separator and stripper pressures can be affected.

o1-preview provides a more plausible explanation by hypothesizing that a decrease in the D feed rate (Stream 2) is the root cause. It attributes the reduction in reactor pressure to a shortage of D, which affects two of the reactions and leads to the accumulation of unreacted gaseous reactants and thus increased reactor pressure. The increase in reactor pressure propagates downstream, raising the pressures in the separator and stripper. The reduced availability of D also explains the reduction in the production of byproduct F through one of the reactions, accounting for the observed decrease in F in the purge stream. In summary, o1-preview concludes with the same root cause as GPT-4o (decrease in D feed) but is able to provide an explanation with stronger causal links.

The failure of both models underscores a limitation in the general reasoning prompt setting: the inability of the PCA-selected features to capture the subtle dynamics of reaction kinetics changes. Both models rely on indirect indicators (e.g., feed changes or pressure increases) to construct plausible but incorrect explanations. This highlights the importance of integrating richer domain knowledge or alternative feature selection methods to better represent complex root causes, particularly those involving gradual or non-obvious changes like reaction kinetics drift.

\section{Conclusions}\label{sec:conclusions}
In this paper, we present \emph{FaultExplainer}, a tool for diagnosing and explaining faults in the Tennessee Eastman Process (TEP). \emph{FaultExplainer} combines real-time sensor data visualization, $T^2$ fault detection and feature contribution analysis, and an interactive chat interface powered by large language models (LLMs). Users can monitor sensor data, introduce faults into the process, and visualize their progression through $T^2$ plots. The interactive chat interface provides grounded explanations based on process data and allows users to ask qualitative questions about fault causes and mitigation strategies, making advanced diagnostic methods more accessible to engineers and operators.

\emph{FaultExplainer} offers several advantages. First, its integration of real-time sensor data and fault detection enhances situational awareness, enabling users to observe and understand deviations in process behavior as they occur. Second, the use of LLMs allows the system to provide explanations that are both grounded in process data and tailored to the user's queries, bridging the gap between raw statistical analysis and actionable insights. Third, its ability to accommodate both predefined root causes and general reasoning makes it flexible for diagnosing faults that may not have been previously encountered.

Despite these strengths, \emph{FaultExplainer} has limitations. The tool relies on Principal Component Analysis (PCA) for feature selection, which may fail to capture subtle or indirect relationships between process variables and root causes, as seen in certain faults where neither model correctly identified the root cause. Furthermore, while LLMs can generate plausible explanations, they sometimes produce overly generalized or hallucinated responses, especially when feature data do not directly relate to the fault mechanism.

Future improvements to \emph{FaultExplainer} could address these limitations. First, integrating advanced feature selection techniques, such as domain-informed feature engineering or causal inference, could enhance the model's ability to detect faults linked to complex root causes. Second, incorporating fine-tuned LLMs trained on chemical process data could improve the specificity and reliability of explanations. These enhancements would further solidify \emph{FaultExplainer}'s role as a powerful tool for real-time fault diagnosis and explanation in industrial processes.

\section{Acknowledgements}
The authors gratefully acknowledge the Purdue Process Safety and Assurance Center (P2SAC) for providing financial support for this work.


 \bibliographystyle{elsarticle-harv}

\end{document}